# Real Time Multi-Class Object Detection and Recognition Using Vision Augmentation Algorithm

Al-Akhir Nayan[1], Joyeta Saha[1], Ahamad Nokib Mozumder[1], Khan Raqib Mahmud[2], Abul Kalam Al Azad[3]

[1]*Lecturer, Department of Computer Science & Engineering, European University of Bangladesh, Dhaka, Bangladesh*
[2]*Lecturer, Department of Computer Science & Engineering, University of Liberal Arts Bangladesh, Dhaka, Bangladesh*
[3]*Associate Professor, Department of Computer Science & Engineering, University of Liberal Arts Bangladesh, Dhaka, Bangladesh*

*(Corresponding author's e-mail: asquiren@gmail.com)*

**Abstract**

*The aim of this research is to detect small objects with low resolution and noise. The existing real time object detection algorithm is based on the deep neural network of convolution need to perform multilevel convolution and pooling operations on the entire image to extract a deep semantic characteristic of the image. The detection models perform better for large objects. The features of existing models do not fully represent the essential features of small objects after repeated convolution operations. We have introduced a novel real time detection algorithm which employs upsampling and skip connection to extract multiscale features at different convolution levels in a learning task resulting a remarkable performance in detecting small objects. The detection precision of the model is shown to be higher and faster than that of the state-of-the-art models.*

## 1. Introduction

Human look at an image and instantly identify what objects are within the image, wherever they are and how they act. The human sensory system is quick and accurate that allows us to perform advanced tasks like driving under very unfavorable situation. Fast, accurate and developed algorithms for object detection would enable computers to perform human-like smart tasks, such as driving vehicles in any weather conditions without specific sensors, or empowering assistive gadgets to pass real-time scene visual data to human clients, and thus unlock the potential of responsive robotic systems for general purposes [1-4].

Object detection is the process of finding real-world objects like faces, bicycles, buildings, etc., from images or videos [5]. Current object detection algorithms such as You Only Look Once (YOLO) [6-7], Single Shot Multibox Detector (SSD) [8], Mask Region-Convolutional Neural Network (Mask RCNN) [9], Fast Region-Convolutional Neural Network (Fast RCNN) [10], Faster Region-Convolutional Neural Network (Faster RCNN) [11], RetinaNet [12]. , Repurpose Classifers [13-14] for detection. For the detection of an object these networks use classifiers for that object and evaluate it in a test image at different locations and scales. A sliding window approach is used by Systems like Deformable Parts Models (DPM) [15-16] where the classifier runs across the entire image at evenly spaced locations.

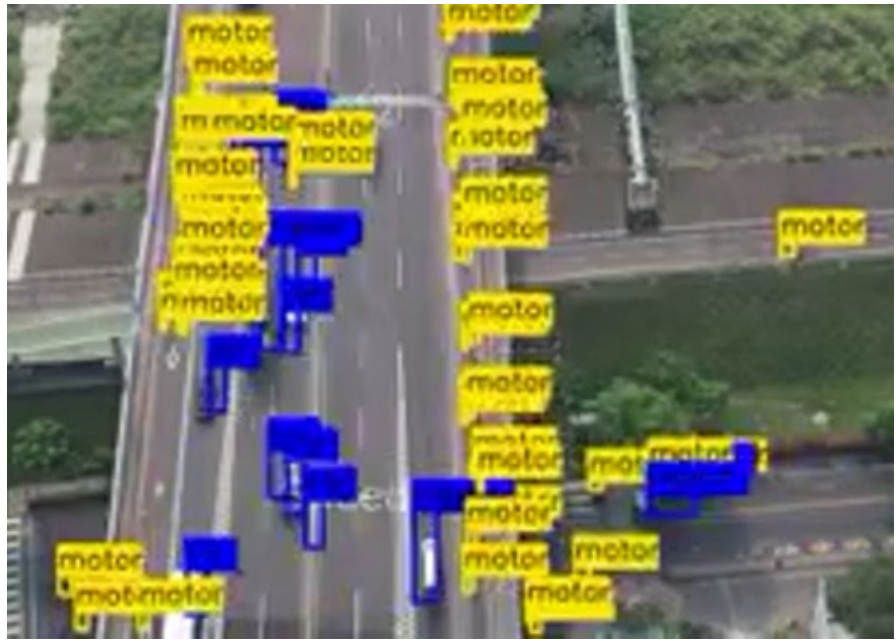

**Figure 1. Real Time Small Object Detection**

The existing object detection literature focuses on detecting a large object that spans a large part of the image. The problem of finding a small piece of an image is largely overlooked [17-18]. The state-of-art algorithm for detecting small object offers unsatisfactory performance. due to the low resolution of the feature map resulting from convolution, and as a result the small object features get too small to be detectable. Our study targets this problem and offers solution to bridge the gap in the current image processing venture. To evaluate the performance of small object detection, we first create a benchmark data set tailored to the problem of small object detection. Small object detection implies detecting small objects with low resolution and dominated by the environment. Small object detection has, in recent times, become very popular in the field of research because it can be helpful in satellite remote sensing, navigation, driving autonomous car and space observation.

In case of real time object detection, high rapidity is very important. And it is a big challenge for the researchers to maintain good accuracy with high rapidity in performance. We refer to YOLOV3 (Figure 2) model because of its several benefits like fast speed and good detection accuracy. The model was modified as per our demand and the modified model has more benefits over traditional models.

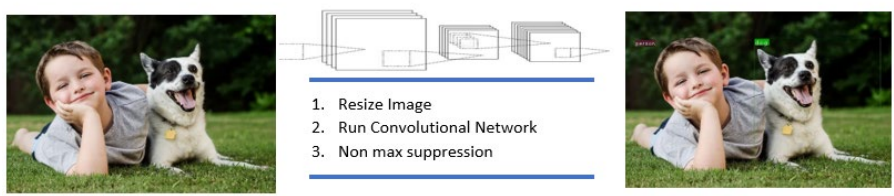

**Figure 2. The YOLO Detection System.**

Important features of our algorithm are:
1. Feature resolution is made high by increasing the number of upsampling layers. Upsampling layers increase the lower resolution and the small objects do not lose its properties.
2. The speed of the model has been maintained 43 FPS using Nvidia GPU. This indicates that real time video steaming can be processed less than 20 milliseconds of latency.
3. The number of shortcut layers and the convolutional layers has been modified. The modified model was trained by increasing the number of shortcut and decreasing convolutional layers. From training of 130000 images, it achieved a training accuracy of 0.972.
4. Since small object purposed dataset is not available, a huge dataset of 130000 images contained 20000 small object images was made. The model was trained through the dataset and a good accuracy was maintained.

All of our training and testing code has been uploaded to github. Even our pretrained model is available in there. At present we have kept the repository private because of paper review process.

## 2. Related Works

Real time object detection has become a highly trending field of research in computer vision, given the wide scope of real-life applications. Typically, to handle image resolution issues, max-pooling is implemented in vision models, however, most of the models skip small objects during detection tasks. Driven by the motivation to address this shortcoming in detecting objects with relatively smaller spatial extent against the multimodal and busy backdrop, the study proposes a modification to the YOLOv3, which conducts detection and classification of objects by generating bounding boxes on images. Furthermore, as different types of classifier influence the accuracy of object detection, the classifier has been adjusted according to different types of detected objects to get good performance.

Deep Learning has been used heavily since 2012 [few references] to achieve the best possible classification accuracy in the ImageNet competition, and eventually deep learning has became a major computational technique for detecting objects. The deep learning based object detection model is divided into two categories: firstly, regional proposals [19-20], such as RCNN [9], SSP-Net [21], Fast-RCNN [10], Faster-RCNN [11] and R-FCN [22]. The other approach does not use suggestions for regions but advises to detect the objects directly, such as, detection done by YOLO [6-7] and SSD [8].

The model selects Region of Interest (RoI) for the regional proposal method during first detection; i.e., selective search [23], edge box [24], or RPN [25] are used to produce multiple RoIs. The model then extracts features by CNN for each RoI, and classifies objects by classifier, and finally locates detected objects. RCNN [9] uses selective search [26] to generate around 2,000 RoIs for each picture and then extract and identify the 2000 RoIs convolution characteristics. Since these RoIs have many overlapping elements, the large number of repeated calculations contributes to inefficient detection. SSP-net [21] and Fast-RCNN [10] propose a specific RoI feature for this problem. The methods extract only one CNN function from the entire original image, and the RoI pooling operation extracts the feature of each RoI independently of the CNN function. Consequently, the calculation number for the extraction function of each RoI is shared. This method reduces the CNN operation needed 2000 times in RCNN to one operation in CNN, and thus increases the computation speed considerably.

The other form does not have a proposed region for object detection. YOLO [6-7] breaks the entire original picture into the SxS grid. If the center of an object falls within a cell, the object is identified by the corresponding cell and the confidence score for each cell is designated. The score indicates the possibility of the target being in the boundary box as well as the precision of Intersection over Union (IoU). YOLO does not use region proposal but translates operations directly over the entire image, so speed is faster than Faster-RCNN, but the accuracy is comparatively. SSD [8] also uses a single neural convolution network to transform the image, and predicts a series of boundary boxes of different sizes and length and width ratios for each point. During the test phase, the network predicts the probability of each object class in each boundary box and changes the boundary box to the shape of the object. G-CNN [26-27] sees object detection as an issue moving the detection box from a fixed grid to an individual box. The model first divides the entire image into different scales to obtain the initial bounding box and extracts the characteristics from the entire image through conversion process. The feature image surrounded by an initial bounding box is then modified to a fixed size image by the Fast-RCNN method. Eventually, more accurate bounding box is availed by regression process.

In short, between the two types of object detection methods for the current mainstream, the first one is more accurate but the speed is slower. The accuracy of the second one is slightly worse, but having advantage of being faster. Regardless of the way the object is identified, the feature extraction uses multilayer convolution process, which can provide the rich abstract object functionality for the target object. But this approach leads to a reduction in detection accuracy for small target objects, since the features collected by the method are few and cannot fully represent the object's characteristics.

## 3. Research Method

### 3.1 The Model

YOLOv3 is the third object detection algorithm within the YOLO (You Only Look Once) family. It is a feature learning-based network comprising 75 convolutional layers. There is no fully-connection layer, there is no form of pooling mechanism either. Stride is used for down-sampling and moving the feature map. A ResNet-alike [28] and FPN-alike structures [29] are also a key to improving its accuracy. Figure 3 below provides full details of the YOLOv3 network.

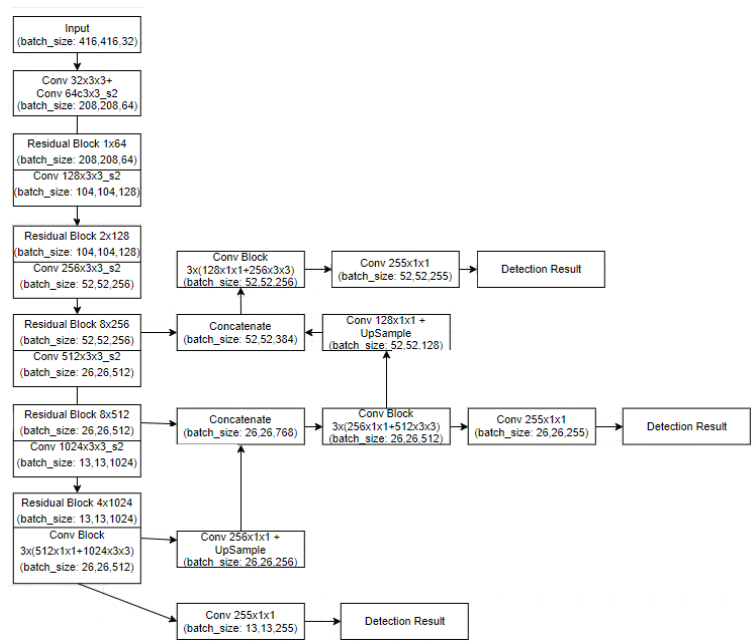

**Figure 3. YOLOv3 Network Architecture**

The input images are used during training to predict 3D tensors which correspond to three scales (Figure 4); the three scales aim at detecting items of different sizes. If the ground-truth bounding box in the middle of the object falls in a certain grid cell, this grid cell is responsible for determining the boundary box of the object. The corresponding object score is assigned "1" for this grid cell, and "0" for others. Three preceding boxes of different sizes are assigned for each grid cell.

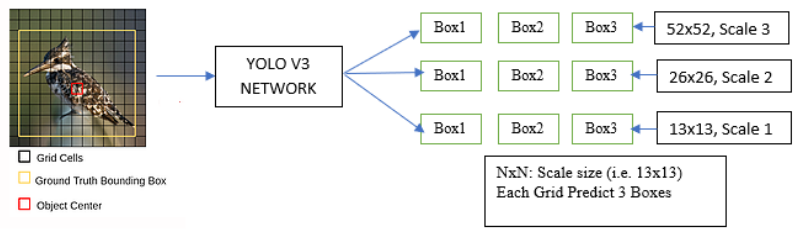

**Figure 4. YOLOv3 Process Flow**

K-mean clustering is used to classify the total bounding boxes from dataset to 9 clusters. This selects the cluster sizes in 3x3 scales. This prior information is helpful for the network to learn compute box coordinate precisely because intuitively bad choice of box size makes it harder and longer for the network to learn.

### 3.2 Modified Model Architecture

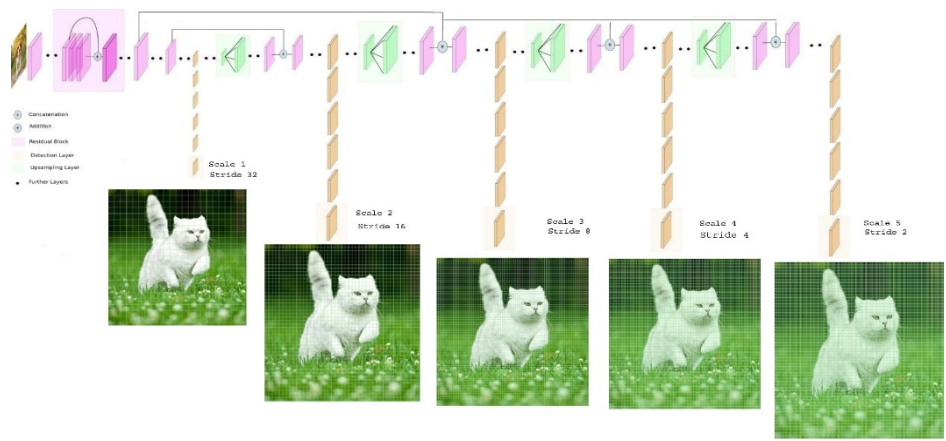

**Figure 5. Proposed Model Architecture**

The figure 5 shows the architecture of the modified model. It consists of 60 convolutional layers, 20 skip connections and 6 upsampling layers. The number of convolutional layers has been decreased from 75 to 60 and upsampling layers have been increased from 2 to 6 in order to boost training. Besides, the number of upsampling layers has settled to 20. Upsampling layers are being used for recreating image resolution. While due to convolution process the resolution of images decreases, upsampling layers recreate the resolution so that the information does not get lose by the learning process of the model. Max pooling has not been used in model implementation. Researchers argue that max pooling is responsible for information loss. It has kept unaltered to ensure better output. Conv 1x1 has kept unaltered as it does not make change to the size of image in the output. The model estimates prediction through bounding boxes.

### 3.3 Improvement to Anchor Box

The number of predefined bounding boxes, also known as anchor, has been increased in order to facilitating the detection of small objects among the bigger objects which tend to overwhelm the visual scene. The height and width of the bounding boxes were also modified accordingly. To account for the five detectors per cell as we have, five anchors are assigned. The task allocation of the detectors are following: small objects are collected from detector 1, slightly larger objects from detector 2, long but flat objects from detector 3, tall but thin objects from detector 4, and large objects from detector 5.

### 3.4 Loss Function

We introduced modification to the YOLO v3 loss function and the different functional parts of the proposed loss function are following:

Localization:

$$\lambda_{coord} \sum_{i=0}^{S^2} \sum_{j=0}^{B} 1_{ij}^{obj} [(x_i - \hat{x}_i)^2 + (y_i - \hat{y}_i)^2] +$$

$$\lambda_{coord} \sum_{i=0}^{S^2} \sum_{j=0}^{B} 1_{ij}^{obj} \left[ (\sqrt{w_i} - \sqrt{\hat{w}_i})^2 + (\sqrt{h_i} - \sqrt{\hat{h}_i})^2 \right]$$

Confidence:

$$+ \sum_{i=0}^{S^2} \sum_{j=0}^{B} 1_{ij}^{obj} (c_i - \hat{c}_i)^2 + \lambda_{noobj} \sum_{i=0}^{S^2} \sum_{j=0}^{B} 1_{ij}^{noobj} (c_i - \hat{c}_i)^2$$

Classification:

$$+ \sum_{i=0}^{S^2} 1_{ij}^{obj} \sum_{c \in classes} (p_i(c) - \hat{p}_i(c))^2$$

Here, the localization part decides the existence of objects under bounding box area, the confidence part decides having no objects and the classification part predicts objects. Here is objectness score, $b_x$, $b_y$, $b_w$, $b_h$ are the center co-ordinate, $p_1, p_2, p_3$ are class confidence scores, box co-ordinates are $t_x$, $t_y$, $t_w$, $t_h$ and $p_w$, $p_h$ are anchor dimensions.

### 3.5 Modification to Scales for Detection

The model makes prediction through 5 different scales; the detection layer is used to detect feature maps of five different sizes, each with steps 32, 16, 8, 4, 2. This means that we detect on scales 13 x 13, 26x 26, 52x 52, 104x 104 and 208x 208 with an input of 416x 416. The network initiates downsampling the input image to the first layer of detection, where the feature maps of a stride 32 layer are used to detect it. Layers are upsampled by stride 2 and concatenated with feature maps of previous layers with the identical sizes of feature map. Another detection is currently created with stride 16 on the layers. The same upsampling procedure is repeated and a final detection is carried out at the stride 16, 8, 4 and 2 layers. Each cell predicts 5 anchors at each scale and the total number of anchors used is 25. Upsampling can help the network recreating the resolution so the network learns efficiently and correctly both the small and big objects.

For an image size 416 x 416, the model predicts ((208 x 208) + (104 x 104) + (52 x 52) + (26 x 26) + 13 x 13)) x 5 bounding boxes. This is staggeringly high number of bounding boxes with heavily weighing on the computation time and resource. Thresholding has been used to minimize it; the values below a certain assigned limit curtails the related anchors.

### 3.6 Model Parameters

The model uses Leaky Relu activation and predict through 5 anchor boxes. The model parameters have been listed in the Table 1.

**Table 1. Parameters Used in Model**

| Name_of_Parameters | Symbol / values |
|---|---|
| Stride | 32,16,8,4,2 |
| Number_of_anchor_boxes | 5 |
| Filter | 32,64,128,256 |
| Size | 5 |
| Random | 1 |
| Truth ithreshold | 1 |
| Ignore threshold | 0.7 |
| Jitter | 0.3 |
| Num | 9 |
| Classes | 80 |
| Mask | 0,1,2 |
| Route layers | -4, -1, 61 |
| Shortcut from | -3 |
| Pad | 1 |

### 3.7 Dataset

A dataset was prepared to train the modified YOLO v3 model. Images were gathered from different sources like Sun Database, Google Photos and captured images. Approximately 130000 images with 115 categories were collected. The images containing both small and big objects were used for training the model. Before training, the images were preprocessed; rescaling, standardization, normalization, binarization were conducted for preprocessing. Gaussian distribution was used to standardize data, while Gaussian blur was employed to remove noise from images. The configuration of our prepared dataset are presented in the table 2 and 3.

**Table 2. Dataset configuration**

| | | Properties |
|---|---|---|
| | Number_of_object Classes | 115 |
| **Number of Images** | Training | 112300 |
| | Validation | 6055 |
| | Testing | 10991 |
| | Regulation of Image (Average) | 469x387_pixels |
| | Per image object classes average | 1.221 |
| | Object scale in average | 0.215 |

**Table 3. Small Object Dataset**

| Category | Number_of_images | Number_of_objects |
|---|---|---|
| Mouse | 282 | 358 |
| Mobile | 265 | 332 |
| Outlet | 305 | 477 |
| Faucet | 423 | 515 |
| Clock | 387 | 422 |
| Toilet_paper | 245 | 289 |
| Bottle | 209 | 371 |
| Plate | 353 | 575 |

### 3.8 Hardware & Software Tools

We used machine that contained intel core i5 processor with 2 GB intel graphics memory + 2GB NVidia GeForce 840m graphics memory. Device memory configuration: 240 GB SSD and 12 GB RAM. As for the software, python 3.6, CUDA, Nvidia CUDA toolkit, pyTorch, OpenCV, TensorFlow, Matplotlib were used implementation and experiments.

## 4. Results And Discussion

The model used two steps training method. Initially, the model was trained and it generated a weight file. The weight file, then, used to obtain the final detection result. The weight file was generated through 8,000 iterations. Iteration indicates the number of passes using batch. One pass is the sum of one forward pass and one backward pass. For an example, say for 128000 training examples with applied batch size of 16, the number of iterations will be 8000. The model scored 0.912 mAP at the time of training. And the test accuracy was maintained 0.829 (Figure 6).

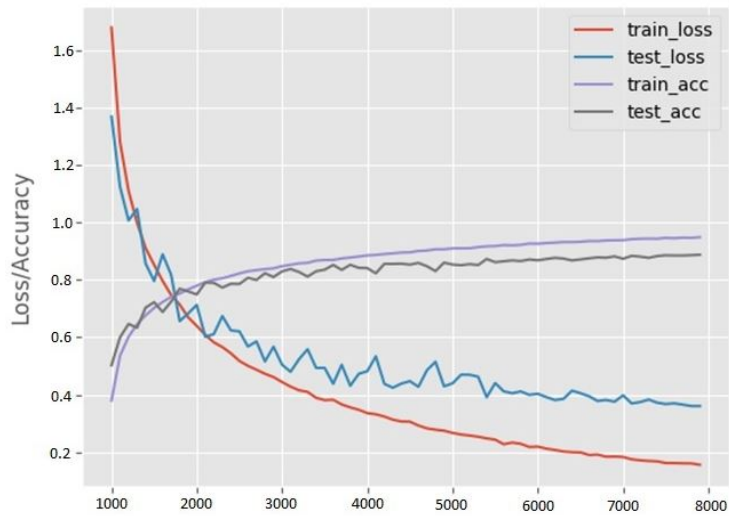

**Figure 6. Training and Testing Loss/Accuracy**

### 4.1 Accuracy & Speed Comparison with Other Models

The paper compares the modified model to Faster-RCNN, Retina Net, YoloV2, YoloV3, SSD and R-FCN. The column chat mentioned below will give information about the modified model and other related models pertaining accuracy.

**Table 4. Accuracy and Rapidity Comparison**

| Model_Name | Accuracy_(mAP) | Rapidity_(FPS) |
|---|---|---|
| RetinaNet | 57.5 | 5 |
| R-FCN | 51.9 | 12 |
| SSD | 46.4 | 19 |
| Yolo_V2 | 48.1 | 40 |
| Yolo_V3 | 51.5 | 45 |
| Modified _Model | 59.6 | 43 |

The modified model evidently has scored higher in accuracy than traditional YOLO and related models (Table 4). The use of multi-scaling feature, modified upsampling and shortcut layers, modified anchor boxes and convolution layers rendered the training performance more accurate compared to other models.

The modified model was designed such a way that it could prevent data or information loss. The model carries a lot of information because of using multi-scaling and 5 predefined bounding boxes for detection and classification. A huge number of information was needed to analyze by the model to prepare output. This compromised a bit the speed of the model process. Nevertheless, the model still maintains a good processing speed which can detect small objects from a real-time video image. The model can process 43 frame per second (Table 4) which indicates that output can be delivered within 23 milliseconds delay.

### 4.2 Comparison of Mean Average Precision with Faster RCNN

During implementation and experiments, the loss in the task performance has been decreased by the increasing the number of processing iterations. Here we measured loss incurred during processing 8 particular small objects and compared with the estimated loss form similar working model (Faster RCNN). The model achieved an accuracy of 0.748 illustrated in Table 4. Our model scored a loss of 0.252, which is lower than Faster RCNN.

Mean average precision is a metric for measuring the accuracy of object detectors. It is the mean average accuracy of various recall values. The accuracy scored by the model has been listed in table 4.3. Here the accuracy has been compared with that of Faster RCNN. Faster RCNN has introduced small object detection multi-scale feature but fail to work in real time because of its low speed and poor accuracy.

**Table 5. Mean Accuracy Precision our model vs Faster RCNN [11]**

|  | Faster RCNN | Modified Model |
|---|---|---|
| mAP | 0.589 | 0.748 |
| Mouse | 0.687 | 0.941 |
| Mobile | 0.6 | 0.731 |
| Outlet | 0.641 | 0.785 |
| Faucet | 0.506 | 0.693 |
| Clock | 0.585 | 0.712 |
| Toilet_Paper | 0.482 | 0.695 |
| Bottle | 0.806 | 0.79 |
| Plate | 0.402 | 0.635 |

### 4.3 Detection Comparison

Accuracy of the detection is stable when the number of YOLO network iterations is 8000 and the number of detection network iterations is 200000 as observed from the experiments. In this study, the comparison is conducted against Faster-RCNN, because it was used to detect small objects. It is found that the proposed modified model is more accurate than others for all object types.

Remote sensing images are also detected in the real environment. The remote sensing image dataset comes from the Google map and UAV (unmanned aerial vehicle) [30, 31] photographs the field transmission line insulators. Because the images in the real environment have the characteristics of changing light, complex background and incomplete objects, so in the construction of the dataset for the study considered all the potential special cases. Experiments show that in real environments our proposed detection model produces better results in detecting small objects. The object detection section renderings are shown in Figure 7.

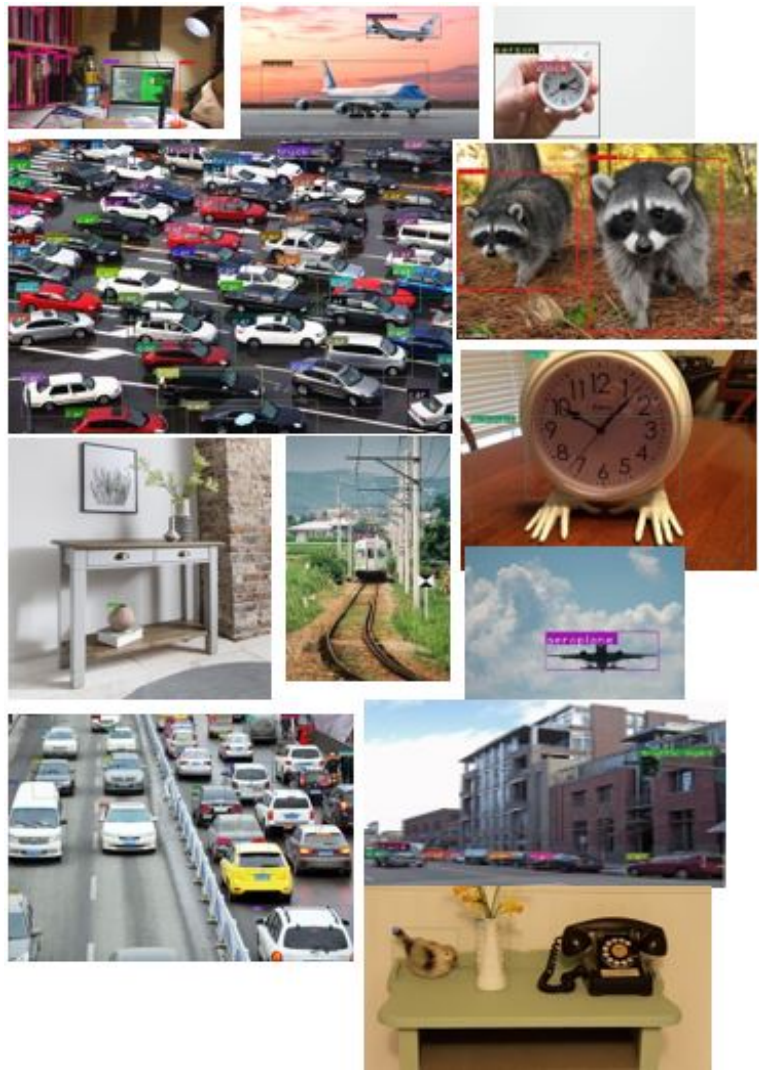

**Figure 7. Object Detection from Image**

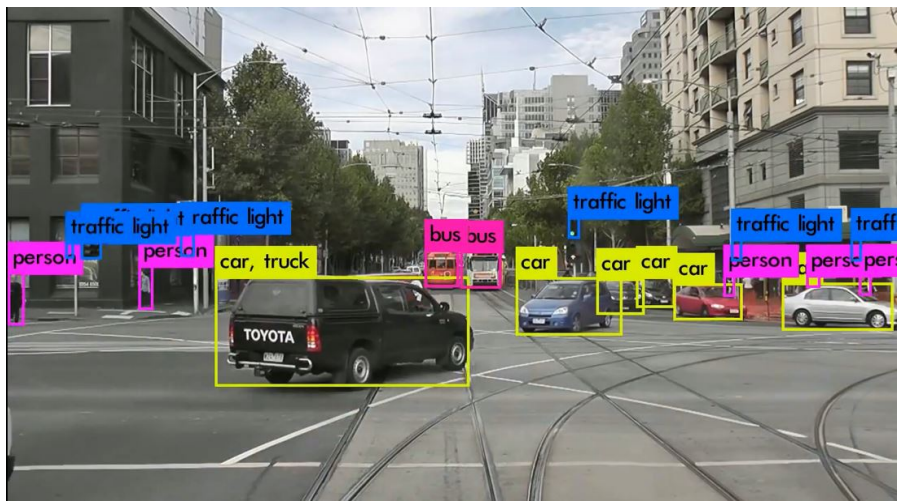

**Figure 8. Real Time object Detection**

Figure 8 illustrates and example of detection of objects using the proposed model from real time video. Evidently, small objects, like the traffic, person, and cars are being detected accurately and promptly. Though the objects are far from the focus area and looking small, the model can pick them up and identify the classes from the prevalent busy environment.

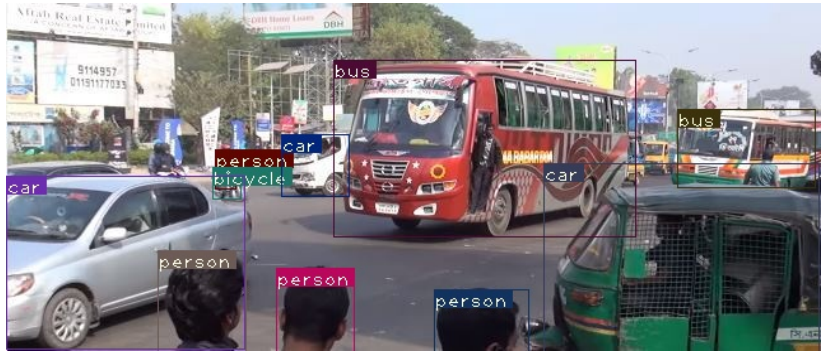

**Figure 9. Real Time Object Detection in Bangladesh**

Another example in figure 9 shows the real time object detection from a busy traffic in Bangladesh. In this example the model detects persons although they are partially occupying the inspecting frame. Performances such as these signify the accuracy of the proposed model in detecting objects with considerable spatial dimensional constraints.

## 5. Conclusion

Small objects are very difficult to detect in real time, due to their lower resolution and greater influence on the environment. Existing detection models based on a deep neural network cannot detect small objects in real time because the properties of objects are extracted by multiple convolutions. Further, due to pooling activities information is lost. Our model not only retains the integrity of the function of the large object, but also preserves the full details of the small and large objects by extracting the multi-scale image by employing skipping connections and upsampling layers in a standard real time detection model. It provides high speed and accurate detection of, both small and large objects in complex environmental setting in real time. In the modified model, various spatial scaling characteristics were introduced resulting in the images being divided into numerous smaller grids. Consequently, this facilitates the detection of smaller objects as they spatially traverse the smaller grids and get easily detected. Additionally the adjustments related to anchor boxes' height and width also aid identification. We maintain our modifications to the detection algorithm are general and applicable beyond the test model we used in this study to wider varieties of real time models, and should be able to be used in wider scope of applications in object detection as the modifications empower the models to perform precise detection of objects of smaller dimension.